\def\year{2022}\relax
\documentclass[letterpaper]{article} 
\usepackage{aaai22}  
\usepackage{times}  
\usepackage{helvet}  
\usepackage{courier}  
\usepackage[hyphens]{url}  
\usepackage{graphicx} 
\usepackage{natbib}  
\usepackage{tablefootnote}
\usepackage{multirow}
\usepackage{graphicx}
\usepackage[table,xcdraw]{xcolor}
\usepackage{caption} 
\DeclareCaptionStyle{ruled}{labelfont=normalfont,labelsep=colon,strut=off} 
\usepackage{algorithm}
\usepackage{algorithmic}
\usepackage[basic,italic,defaultimath,nohbar,defaultmathsizes]{mathastext}
\usepackage{CJKutf8}

\usepackage{CJK}

\usepackage{subcaption}
\usepackage{verbatim}
\usepackage{colortbl}
\usepackage{graphicx}
\definecolor{dGray}{gray}{.6}
\definecolor{mGray}{gray}{.9}
\definecolor{lGray}{gray}{.94}

\usepackage{xcolor}

\newcommand{\answerTODO}[1]{\textcolor{red}{#1}}

\usepackage{newfloat}
\usepackage{listings}
\floatstyle{ruled}
\newfloat{listing}{tb}{lst}{}
\floatname{listing}{Listing}
\title{Examining Temporal Bias in Abusive Language Detection}
 \author{
    Mali Jin,
    Yida Mu,
    Diana Maynard,
    Kalina Bontcheva   
}
\affiliations{
    \textsuperscript{\rm }Department of Computer Science, The University of Sheffield, UK\\

    {\{m.jin, y.mu, d.maynard, k.bontcheva\}@sheffield.ac.uk} \\
}

\begin{document}

\maketitle

\begin{abstract}

The use of abusive language online has become an increasingly pervasive problem that damages both individuals and society, with effects ranging from psychological harm right through to escalation to real-life violence and even death. Machine learning models have been developed to automatically detect abusive language, but these models can suffer from temporal bias, the phenomenon in which topics, language use or social norms change over time. This study aims to investigate the nature and impact of temporal bias in abusive language detection across various languages and explore mitigation methods. We evaluate the performance of models on abusive data sets from different time periods. Our results demonstrate that temporal bias is a significant challenge for abusive language detection, with models trained on historical data showing a significant drop in performance over time. We also present an extensive linguistic analysis of these abusive data sets from a diachronic perspective, aiming to explore the reasons for language evolution and performance decline. This study sheds light on the pervasive issue of temporal bias in abusive language detection across languages, offering crucial insights into language evolution and temporal bias mitigation.

\end{abstract}

\section{Introduction}

The increasing use of social media platforms has given rise to a pervasive problem of online abusive language, which can cause harm to individuals and lead to societal polarization. In recent years, researchers have developed a huge variety of machine learning models that can automatically detect abusive language \citep{mishra2019abusive,aurpa2022abusive,das2023transfer,alrashidi2023abusive}. However, these models may be subject to temporal bias, which can lead to a decrease in the accuracy of abusive language detection models, potentially allowing abusive language to be undetected or falsely detected.

Temporal bias arises from differences in populations and behaviors over time \citep{olteanu2019social}. In natural language processing (NLP), it can result from various issues. Temporal concept drift refers to the problem of language evolving over time \citep{zhao2022impact}. Languages change as new meanings develop for existing words and new words and topics come into use over time. Models trained on data from an earlier period can perform worse on chronologically newer data as they are unable to recognize new topics or linguistic features \citep{lukes2018sentiment,vidgen2019challenges,mu2023examining}. Previous work has examined temporal bias in various tasks such as named entity recognition \citep{derczynski2016broad}, sentiment analysis \citep{lukes2018sentiment} and rumour detection \citep{mu2023s}.

In online abuse detection, words and expressions considered acceptable in the past may have an abusive or offensive connotation now due to the changing language or societal norms \citep{wich2022bias,mcgillivray2022leveraging}. Also, temporal bias occurs when the abusive content fluctuates based on the latest trends, popular topics or breaking news. As the online discussion evolves with new development, certain topics and forms of abuse might gain prominence while others become less prevalent. For example, in 2020 a fraudulently altered video was circulated on Twitter purporting to show Al Jazeera journalist Ghada Oueiss naked in a jacuzzi, as part of an orchestrated attack designed to discredit her \citep{posetti2021chilling}. The video and other photos were distributed with messages alleging she was an alcoholic, drug-addicted prostitute, which engendered in turn a large number of hateful messages connected with the alleged jacuzzi incident, a topic not typically associated with abuse. 


Previous work identified temporal bias in an Italian hate speech data set associated with immigrants \citep{florio2020time}. However, they have yet to explore temporal factors affecting predictive performance from a multilingual perspective. In this paper, we explore temporal bias in 5 different abusive data sets that span varying time periods, in 4 languages (English, Spanish, Italian, and Chinese). Specifically, we investigate the following core research questions:

\begin{itemize}
    
    \item  \textit{RQ1:} How does the magnitude of temporal bias vary across different data sets such as language, time span and collection methods?

    \item \textit{RQ2:} What type of language evolution causes the temporal bias in our data sets and how?

    \item  \textit{RQ3:} Could domain adaptation models, large language models (LLMs) or a more robust data set help to mitigate the temporal bias in abusive language detection?
    
\end{itemize}

To answer these questions, we compare the predictive performance between random and chronological data splits across data sets in different languages and with different temporal coverage. We also experiment with different transformer-based pre-trained language models (PLMs) using the original data set and a filtered data set. Finally, we present an in-depth analysis to investigate the factors for performance degradation.






\section{Related Work}

\subsection{Bias in NLP}
Bias refers to the presence of systematic and unfair favouritism or prejudice. In various contexts, bias can manifest as a skewed representation or inaccurate judgments that unfairly advantage or disadvantage certain individuals or groups \citep{garrido2021survey}. Bias can arise from various sources such as data selection, annotation processes, models and research design. These biases can potentially lead to unfair or discriminatory outcomes through NLP applications \citep{hovy2021five}. For instance, biased language models might generate discriminatory content or fail to accurately understand and respond to underrepresented languages. Consequently, addressing and mitigating bias in NLP has become a critical research endeavour. Researchers are exploring techniques to measure and mitigate bias across diverse domains and languages \citep{sun2019mitigating,font2019equalizing,zueva2020reducing,czarnowska2021quantifying}. Common debiasing methods include data reweighing and resampling, debiasing word
embeddings, counterfactual data augmentation and bias fine-tuning \citep{kamiran2012data,zhao2018gender,park2018reducing}.

\subsection{Bias in Abusive Language Detection}
 Previous work has focused on identifying and mitigating different forms of social bias in abusive language detection, such as gender bias \citep{park2018reducing}, dialect bias (e.g. African-Americans English) \citep{davidson2019racial,sap-etal-2019-risk, davidson2020examining,zhou2021challenges} and different forms of identity bias (e.g. transgender, black) \citep{dixon2018measuring,zueva2020reducing}. Moreover, \citet{elsafoury2022sos} measured systematic offensive stereotyping bias (i.e., associating slurs or profane terms with specific groups of people, especially marginalized people) in different word embeddings. 


However, little attention has been paid to temporal bias in abusive language detection. One exception is the work of \citet{florio2020time}, who identified temporal bias in an Italian hate speech data set associated with immigrants. They investigated the impact of data size and time spans on temporal robustness by using two strategies, namely a sliding window model and an incremental model. Their results showed that adding training data temporally closer to the testing set greatly improved the performance but simply increasing the size of training data did not lead to performance improvement. Also, they found that offensive language in online contexts experienced rapid changes in topics over different time periods. Moreover, \citet{mcgillivray2022leveraging} made use of time-dependent lexical features to detect abusive language effectively by training on smaller and older data. To facilitate this, they obtained a list of words for semantic change (i.e. acquired or lost an offensive meaning between 2019 and 2020). Their results showed that semantic change impacts abusive language detection and it is feasible to improve the detection by considering this change instead of depending on large labeled data sets. However, both work restricted themselves only to a single data set or a single language and did not explore other languages.


\subsection{Temporal Bias in Classification Tasks}
Temporal bias occurs in classification tasks due to the variation and evolution of data patterns over time. This temporal variation can pose difficulties for machine learning models as patterns learned from one time period may not be applicable in another. Temporal bias was assessed in various classification tasks such as rumour detection \citep{mu2023s}, stance detection \citep{mu2023examining} and multi-label classification tasks related to legislation and biomedicine \citep{chalkidis2022improved}. \citet{mu2023examining} found that domain-adapted pre-trained language models are less sensitive to time and thus are beneficial to temporal gap mitigation; while \citet{chalkidis2022improved} proposed group-robust algorithms to reduce the temporal bias in multi-label classification. Moreover, \citet{alkhalifa2023building} investigated the impact of word representations and machine learning model choice on temporal performance of various classification tasks such as stance detection and sentiment analysis.

\begin{table*}[!t]
\centering
\resizebox{\textwidth}{!}{%
\begin{tabular}{|c|c|c|c|c|c|}
\hline
\textbf{Dataset} & \textbf{Language} & \textbf{Source} & \textbf{Time} & \textbf{Size} & \textbf{Labels} \\ \hline
\citet{waseem2016hateful} & English & Twitter & 07-04-2013 - 06-01-2016 (33 months) & 16,914 & neither, sexism, racism \\ \hline
\citet{founta2018large} & English & Twitter & 30-03-2017 - 08-04-2017 (10 days) & 80,000 & normal, spam, abusive, hateful\\ \hline
\citet{jiang2022swsr} &  Chinese & Weibo & 06-04-2012 - 26-06-2020 (8 years) & 8,969 & sexism, not sexism\\ \hline
\citet{pereira2019detecting} & Spanish & Twitter &04-02-2017 - 22-12-2017 (10 months) & 6,000 & hate speech, not hate speech\\ \hline
\citet{sanguinetti2018italian} & Italian & Twitter & 26-02-2015 - 25-04-2017 (26 months) & 6,928 & hate speech, not hate speech\\ \hline
\end{tabular}%
}
\caption{Data sets details.}
\label{tab:dataDetail}
\end{table*}

\section{Data}

We study two widely used English abusive data sets (\textit{WASEEM} and \textit{FOUNTA}). We also study a Chinese data set (\textit{JIANG}), a Spanish data set (\textit{PEREIRA}), and an Italian data set (\textit{SANGUINETTI}), in order to explore the impact of temporality on different languages. We choose these data sets because the creation time of each post is provided or accessible (via tweet IDs). Details of the data sets are shown in Table \ref{tab:dataDetail}.

\paragraph{WASEEM}\citep{waseem2016hateful} is an English abusive data set focusing on sexism and racism. They collect the tweets by manually searching common terms related to religious, sexual, gender, and ethnic minorities, and by using the public Twitter search API. They combine these two methods to ensure that non-offensive tweets that contain clearly or potentially offensive words are also obtained. The annotations are created by manual experts and then reviewed by an additional gender study expert. We merge the original \textit{sexism} and \textit{racism} labels into a single \textit{abusive} label, and rename the \textit{neither} label as \textit{non-abusive}.

\paragraph{FOUNTA}\citep{founta2018large} is an English data set collected from Twitter containing two types of online abuse expressions: abusive and hateful. They randomly collect and sample the data, using text analysis and machine learning techniques to create the boosted set of tweets which are likely to belong to the two abusive classes. The data is then annotated by crowdsourced workers. Similar to \citet{leonardelli2021agreeing}, we map the four labels in the data set into a binary offensive or non-offensive label. We exclude tweets labeled as \textit{spam}, and merge \textit{abusive} and \textit{hateful} labels into \textit{abusive}. The \textit{normal} label is renamed \textit{non-abusive}.

\paragraph{JIANG}\citep{jiang2022swsr} is a Chinese sexism data set collected from Sina Weibo (a Chinese microblogging platform). They first collect gender-related weibos by searching keywords such as `feminism' and `gender discrimination'. Then they extract the comments that link to these weibos and filter out the comments to produce the final data set, which is annotated by three PhD students.

\paragraph{PEREIRA}\citep{pereira2019detecting} is a Spanish hate speech data set annotated by experts. They randomly collect the data using the Twitter Rest API and filter it using seven dictionaries, where six of them represent different types of hate speech (e.g., race, gender) and the last one contains generic insults.

\paragraph{SANGUINETTI}\citep{sanguinetti2018italian} is an Italian hate speech data set targeting immigrants, Roma and Muslims. They obtain the tweets by selecting a set of neutral keywords related to each target. The data is annotated by a team of both expert and crowdsourced annotators.

\subsection{Data Filtering}
Since three is no time information or tweet content in the FOUNTA and SANGUINETTI datasets, we re-obtain the tweets with their created time using Twitter Academic API based on the provided tweet IDs. Given the provided tweet IDs and related texts in the PEREIRA corpus, we use them directly without re-collecting the data to avoid data loss as Twitter ids are time ordered\footnote{https://developer.twitter.com/en/docs/twitter-ids}, For all data sets, we remove the duplicates and any tweets with no created time information.

\subsection{Data Splits}
We divide the data into training and testing sets using two strategies, namely random splits and chronological splits. The statistics of each data set are shown in Table \ref{t:statistics}. We can see that two of the data sets cover only a short period (FOUNTA contains many tweets but only covers 10 days, while PEREIRA covers 10 months but is fairly small in size) while all the other datasets span several years.

\paragraph{Random Splits}
We randomly split the data sets into training and testing sets and keep class distribution the same as the original data sets. 

\paragraph{Chronological Splits}
We adopt a stratified chronological split strategy following the method in \citet{mu2023s}. We first sort the abusive and non-abusive texts separately in chronological order. Then, we extract the first 70\% of posts from abusive and non-abusive sets separately and combine them as the training set. Similarly, we combine the last 15\% of posts from abusive and non-abusive sets as the testing set. The middle part of the two sets is merged into the validation set. In this way, the distribution of labels in each set is consistent with the original data.

\renewcommand{\arraystretch}{1.0}
\begin{table}[!t]
\center
\resizebox{0.48\textwidth}{!}{
\begin{tabular}{|l|c|c|c|c|}
\hline
\bf Dataset & \bf Training & \bf Validation & \bf Testing &\bf All \\ \hline
WASEEM & 12,214 & 2,156 & 2,536 & 16,906 \\
FOUNTA & 27,368 & 5,683 & 4,830 & 37,881\\
\hline
JIANG & 6,335 & 1,118 & 1,316 & 8,769\\
PEREIRA & 4,335 & 765 & 900 & 6,000 \\
SANGUINETTI & 2,861 & 595 & 506 & 3,962\\
\hline
\end{tabular}}
\caption{Data sets statistics.}
\label{t:statistics}
\end{table}

\section{Predictive Models}

\paragraph{LR}
We use Logistic Regression with bag-of-words using L2 regularization as our baseline (LR).

\paragraph{BERT} (Bidirectional Encoder Representations from Transformers; \citealp{devlin2018bert}) is a transformer-based \citep{vaswani2017attention} language model, which is pre-trained on large corpora, such as the English Wikipedia and the Google Books corpus. During pre-training, it uses a technique called masked language modeling (MLM) where it randomly masks some of the words in the input text, aiming to predict the masked word based on the context \citep{devlin2018bert}.  We fine-tune the BERT model on abusive language detection by adding an output layer with a softmax activation function.

\paragraph{RoBERTa} is an extension of BERT trained on more data with different hyperparameters and has achieved better performance in multiple classification tasks \citep{liu2019roberta}. We fine-tune RoBERTa in a similar way to BERT.

\paragraph{RoBERTa-hate-speech} This domain adaptation model\footnote{\url{https://rb.gy/k5x9t}} is trained on 11 English data sets for hate and toxicity based on the RoBERTa-base model \citep{vidgen2020learning}.



\begin{table*}[!t]
\centering
\begin{tabular}{|c|l|llll|llll|}
\hline
 &
  \multicolumn{1}{c|}{} &
  \multicolumn{4}{c|}{\textbf{WASEEM}} &
  \multicolumn{4}{c|}{\textbf{FOUNTA}} \\ \cline{3-10} 
\multirow{-2}{*}{\textbf{Model}} &
  \multicolumn{1}{c|}{\multirow{-2}{*}{\textbf{Splits}}} &
  \multicolumn{1}{l|}{Acc} &
  \multicolumn{1}{l|}{P} &
  \multicolumn{1}{l|}{R} &
  F1 &
  \multicolumn{1}{l|}{Acc} &
  \multicolumn{1}{l|}{P} &
  \multicolumn{1}{l|}{R} &
  F1 \\ \hline
 &
  \textit{Random Splits} &
  \multicolumn{1}{l|}{81.94} &
  \multicolumn{1}{l|}{79.27} &
  \multicolumn{1}{l|}{79.08} &
  79.18 &
  \multicolumn{1}{l|}{92.54} &
  \multicolumn{1}{l|}{83.66} &
  \multicolumn{1}{l|}{84.69} &
   84.16\\ \cline{2-10} 
\multirow{-1}{*}{\textbf{\begin{tabular}[c]{@{}c@{}}LR\end{tabular}}} &
  \textit{Chronological Splits} &
  \multicolumn{1}{l|}{74.88} &
  \multicolumn{1}{l|}{76.93} &
  \multicolumn{1}{l|}{62.69} &
  63.15 &
  \multicolumn{1}{l|}{93.26} &
  \multicolumn{1}{l|}{85.56} &
  \multicolumn{1}{l|}{85.28} &
  85.42 \\  \cline{2-10}
  &
  \cellcolor{mGray} Performance Drop &
  \multicolumn{1}{l|}{\cellcolor{mGray} 7.06$\downarrow$} &
  \multicolumn{1}{l|}{\cellcolor{mGray} 2.33$\downarrow$} &
  \multicolumn{1}{l|}{\cellcolor{mGray} 16.39$\downarrow$} &
  \cellcolor{mGray} 16.03$\downarrow$ &
  \multicolumn{1}{l|}{\cellcolor{mGray} 0.72$\uparrow$} &
  \multicolumn{1}{l|}{\cellcolor{mGray} 1.90$\uparrow$} &
  \multicolumn{1}{l|}{\cellcolor{mGray} 0.59$\uparrow$} &
  \cellcolor{mGray} \textbf{1.26}$\uparrow$ \\ \hline
  
 &
  \textit{Random Splits} &
  \multicolumn{1}{l|}{85.73} &
  \multicolumn{1}{l|}{84.10} &
  \multicolumn{1}{l|}{82.65} &
  83.26 &
  \multicolumn{1}{l|}{94.95} &
  \multicolumn{1}{l|}{90.98} &
  \multicolumn{1}{l|}{86.43} &
  88.49 \\ \cline{2-10} 
\multirow{-1}{*}{\textbf{\begin{tabular}[c]{@{}c@{}}RoBERTa\end{tabular}}} &
  \textit{Chronological Splits} &
  \multicolumn{1}{l|}{76.77} &
  \multicolumn{1}{l|}{80.54} &
  \multicolumn{1}{l|}{65.20} &
  66.33 &
  \multicolumn{1}{l|}{94.81} &
  \multicolumn{1}{l|}{91.16} &
  \multicolumn{1}{l|}{85.45} &
  87.99 \\  \cline{2-10}
  &
  \cellcolor{mGray} Performance Drop &
  \multicolumn{1}{l|}{\cellcolor{mGray} 8.96$\downarrow$} &
  \multicolumn{1}{l|}{\cellcolor{mGray} 3.56$\downarrow$} &
  \multicolumn{1}{l|}{\cellcolor{mGray} 17.45$\downarrow$} &
  \cellcolor{mGray} 16.93$\downarrow$ &
  \multicolumn{1}{l|}{\cellcolor{mGray} 0.14$\downarrow$} &
  \multicolumn{1}{l|}{\cellcolor{mGray} 0.18$\uparrow$} &
  \multicolumn{1}{l|}{\cellcolor{mGray} 0.98$\downarrow$} &
  \cellcolor{mGray} 0.50$\downarrow$ \\ \hline
 &
  \textit{Random Splits} &
  \multicolumn{1}{l|}{89.20} &
  \multicolumn{1}{l|}{87.50} &
  \multicolumn{1}{l|}{87.82} &
  87.64 &
  \multicolumn{1}{l|}{96.42} &
  \multicolumn{1}{l|}{93.16} &
  \multicolumn{1}{l|}{91.11} &
  92.09 \\ \cline{2-10} 
\multirow{-1}{*}{\textbf{\begin{tabular}[c]{@{}c@{}}RoBERTa-\\hate-speech\end{tabular}}} &
  \textit{Chronological Splits} &
  \multicolumn{1}{l|}{81.58} &
  \multicolumn{1}{l|}{85.99} &
  \multicolumn{1}{l|}{72.21} &
  74.71 &
  \multicolumn{1}{l|}{96.07} &
  \multicolumn{1}{l|}{92.03} &
  \multicolumn{1}{l|}{90.79} &
  91.39 \\  \cline{2-10}
  &
  \cellcolor{mGray} Performance Drop &
  \multicolumn{1}{l|}{\cellcolor{mGray} 7.62$\downarrow$} &
  \multicolumn{1}{l|}{\cellcolor{mGray} 1.51$\downarrow$} &
  \multicolumn{1}{l|}{\cellcolor{mGray} 15.61$\downarrow$} &
  \cellcolor{mGray} \textbf{12.93}$\downarrow$ &
  \multicolumn{1}{l|}{\cellcolor{mGray} 0.35$\downarrow$} &
  \multicolumn{1}{l|}{\cellcolor{mGray} 1.13$\downarrow$} &
  \multicolumn{1}{l|}{\cellcolor{mGray} 0.32$\downarrow$} &
  \cellcolor{mGray} 0.70$\downarrow$ \\ \hline
 &
  \textit{Random Splits} &
  \multicolumn{1}{l|}{64.47} &
  \multicolumn{1}{l|}{68.96} &
  \multicolumn{1}{l|}{70.88} &
  64.26 &
  \multicolumn{1}{l|}{80.43} &
  \multicolumn{1}{l|}{68.11} &
  \multicolumn{1}{l|}{81.93} &
  70.54 \\ \cline{2-10} 
\multirow{-1}{*}{\textbf{\begin{tabular}[c]{@{}c@{}}OA\end{tabular}}} &
  \textit{Chronological Splits} &
  \multicolumn{1}{l|}{72.36} &
  \multicolumn{1}{l|}{72.53} &
  \multicolumn{1}{l|}{75.89} &
  71.48 &
  \multicolumn{1}{l|}{80.75} &
  \multicolumn{1}{l|}{68.24} &
  \multicolumn{1}{l|}{81.83} &
  70.77 \\  \cline{2-10}
  &
  \cellcolor{mGray} Performance Drop &
  \multicolumn{1}{l|}{\cellcolor{mGray} 7.89$\uparrow$} &
  \multicolumn{1}{l|}{\cellcolor{mGray} 3.57$\uparrow$} &
  \multicolumn{1}{l|}{\cellcolor{mGray} 5.01$\uparrow$} &
  \cellcolor{mGray} 7.22$\uparrow$ &
  \multicolumn{1}{l|}{\cellcolor{mGray} 0.32$\uparrow$} &
  \multicolumn{1}{l|}{\cellcolor{mGray} 0.13$\uparrow$} &
  \multicolumn{1}{l|}{\cellcolor{mGray} 0.10$\uparrow$} &
  \cellcolor{mGray} 0.23$\uparrow$ \\ \hline
\end{tabular}%
\caption{Model predictive performance on English data sets using random and chronological splits. The smallest F1 performance drop (or rise) across models is in bold.}
\label{tab:results_english}
\end{table*}

\begin{table*}[!t]
\resizebox{\textwidth}{!}{%
\begin{tabular}{|c|l|llll|llll|llll|}
\hline
 &
  \multicolumn{1}{c|}{} &
  \multicolumn{4}{c|}{\textbf{JIANG}} &
  \multicolumn{4}{c|}{\textbf{PEREIRA}} &
  \multicolumn{4}{c|}{\textbf{SANGUINETTI}} \\ \cline{3-14} 
\multirow{-2}{*}{\textbf{Model}} &
  \multicolumn{1}{c|}{\multirow{-2}{*}{\textbf{Splits}}} &
  \multicolumn{1}{l|}{Acc} &
  \multicolumn{1}{l|}{P} &
  \multicolumn{1}{l|}{R} &
  F1 &
  \multicolumn{1}{l|}{Acc} &
  \multicolumn{1}{l|}{P} &
  \multicolumn{1}{l|}{R} &
  F1 &
  \multicolumn{1}{l|}{Acc} &
  \multicolumn{1}{l|}{P} &
  \multicolumn{1}{l|}{R} &
  F1 \\ \hline
  &
  \textit{Random} &
  \multicolumn{1}{l|}{76.14} &
  \multicolumn{1}{l|}{73.24} &
  \multicolumn{1}{l|}{72.81} &
   73.01&
  \multicolumn{1}{l|}{77.00} &
  \multicolumn{1}{l|}{70.35} &
  \multicolumn{1}{l|}{70.95} &
   70.64&
  \multicolumn{1}{l|}{86.22} &
  \multicolumn{1}{l|}{73.93} &
  \multicolumn{1}{l|}{77.19} &
   75.36\\ \cline{2-14} 
\multirow{-1}{*}{\textbf{LR}} &
  \textit{Chronological} &
  \multicolumn{1}{l|}{71.50} &
  \multicolumn{1}{l|}{68.28} &
  \multicolumn{1}{l|}{68.76} &
   68.49 &
  \multicolumn{1}{l|}{80.67} &
  \multicolumn{1}{l|}{76.08} &
  \multicolumn{1}{l|}{69.86} &
   71.83 & 
  \multicolumn{1}{l|}{85.21} &
  \multicolumn{1}{l|}{71.71} &
  \multicolumn{1}{l|}{71.71} &
   71.71\\ \cline{2-14}
  &
  \cellcolor{mGray} Performance Drop &
  \multicolumn{1}{l|}{\cellcolor{mGray} 4.64$\downarrow$} &
  \multicolumn{1}{l|}{\cellcolor{mGray} 4.96$\downarrow$} &
  \multicolumn{1}{l|}{\cellcolor{mGray} 4.05$\downarrow$} &
  \cellcolor{mGray} 4.52$\downarrow$ &
  \multicolumn{1}{l|}{\cellcolor{mGray} 3.67$\uparrow$} &
  \multicolumn{1}{l|}{\cellcolor{mGray} 5.73$\uparrow$} &
  \multicolumn{1}{l|}{\cellcolor{mGray} 1.09$\downarrow$} &
  \cellcolor{mGray} \textbf{1.19}$\uparrow$ &
  \multicolumn{1}{l|}{\cellcolor{mGray} 1.01$\downarrow$} &
  \multicolumn{1}{l|}{\cellcolor{mGray} 2.22$\downarrow$} &
  \multicolumn{1}{l|}{\cellcolor{mGray} 5.48$\downarrow$} &
  \cellcolor{mGray} \textbf{3.65}$\downarrow$\\ \hline

 &
  \textit{Random} &
  \multicolumn{1}{l|}{80.68} &
  \multicolumn{1}{l|}{78.95} &
  \multicolumn{1}{l|}{76.65} &
  77.52 &
  \multicolumn{1}{l|}{80.67} &
  \multicolumn{1}{l|}{75.30} &
  \multicolumn{1}{l|}{72.31} &
  73.44 &
  \multicolumn{1}{l|}{88.07} &
  \multicolumn{1}{l|}{78.09} &
  \multicolumn{1}{l|}{72.69} &
  74.85 \\ \cline{2-14} 
\multirow{-1}{*}{\textbf{BERT}} &
  \textit{Chronological} &
  \multicolumn{1}{l|}{78.66} &
  \multicolumn{1}{l|}{76.28} &
  \multicolumn{1}{l|}{77.80} &
  76.81 &
  \multicolumn{1}{l|}{82.78} &
  \multicolumn{1}{l|}{83.15} &
  \multicolumn{1}{l|}{69.72} &
  72.67 &
  \multicolumn{1}{l|}{84.87} &
  \multicolumn{1}{l|}{70.22} &
  \multicolumn{1}{l|}{63.08} &
  65.13 \\ \cline{2-14}
  &
  \cellcolor{mGray} Performance Drop &
  \multicolumn{1}{l|}{\cellcolor{mGray} 2.02$\downarrow$} &
  \multicolumn{1}{l|}{\cellcolor{mGray} 2.67$\downarrow$} &
  \multicolumn{1}{l|}{\cellcolor{mGray} 1.15$\uparrow$} &
  \cellcolor{mGray} \textbf{0.71}$\downarrow$ &
  \multicolumn{1}{l|}{\cellcolor{mGray} 2.11$\uparrow$} &
  \multicolumn{1}{l|}{\cellcolor{mGray} 7.85$\uparrow$} &
  \multicolumn{1}{l|}{\cellcolor{mGray} 2.59$\downarrow$} &
  \cellcolor{mGray} 0.77$\downarrow$ &
  \multicolumn{1}{l|}{\cellcolor{mGray} 3.20$\downarrow$} &
  \multicolumn{1}{l|}{\cellcolor{mGray} 7.87$\downarrow$} &
  \multicolumn{1}{l|}{\cellcolor{mGray} 9.61$\downarrow$} &
  \cellcolor{mGray} 9.72$\downarrow$\\ \hline

\end{tabular}
}
\caption{Model predictive performance on a Chinese, Spanish and Italian data set using random and chronological splits. The smallest performance drops (or rise) across models are in bold.}
\label{tab:results_otherLanguage}
\end{table*}

\paragraph{OA} We use
the OpenAssistant (OA) 30B model developed by LAIONAI, which fine-tunes the LLaMA (Large Language Model Meta AI; \citealp{touvron2023llama}) 30B model using the OA dataset. Since the original LLaMA model is not fully open-source, we obtain the xor weights from HuggingFace\footnote{\url{https://rb.gy/qfpc9}} and apply 8-bit quantisation techniques via BitsAndBytes \cite{dettmers20228bit} to decrease the inference memory requirements. We use OA for zero-shot classification where we provide the model with a sequence of texts and a prompt that describes what we want our model to do.

\section{Experimental Setup}


\paragraph{Tweet Pre-Processing}
For all data sets, we replace username mentions and hyperlinks with placeholder tokens, $<$USER$>$ and $<$URL$>$ respectively. For the Chinese data set, we use Jieba\footnote{\url{https://github.com/fxsjy/jieba}}, a Chinese text segmentation, to tokenize the texts.

\paragraph{Hyperparameters}
For all the English data sets, we use RoBERTa-base\footnote{\url{https://huggingface.co/roberta-base}}; for data sets in other languages, we use bert-base-chinese\footnote{\url{https://huggingface.co/bert-base-chinese}}, bert-base-spanish-wwm-cased\footnote{\url{https://rb.gy/br2ys}} and bert-base-italian-cased\footnote{\url{https://huggingface.co/dbmdz/bert-base-italian-cased}} respectively, which are trained on big corpora of the corresponding language based on the BERT-base model.
We fine-tune all models with learning rate $l$ = 3e-6,  $l \in$ \{1e-4, 1e-5, 5e-6, 3e-6, 1e-6, 1e-7\}. The batch size is set to 32 and the maximum sequence length is set to 128. All experiments are performed on a NVIDIA Titan RTX GPU with 24GB memory. We follow the official guidelines\footnote{\url{https://huggingface.co/OpenAssistant}} to run the 30B OA model on a local server with two NVIDIA A100 GPUs.

\paragraph{Training and Evaluation}
We split the data sets into training, validation and testing sets with a ratio of 70:15:15. During training, we choose the model with the smallest validation loss value over 12 epochs. We run all models five times with different random seeds for both random and chronological split strategies. We report predictive performance using the average Accuracy, Precision, Recall and macro-F1 scores. For OA, we only input the prompt (i.e. \textit{identify if the following text is abusive or non-abusive}) and the same testing sets using two data split strategies.

\section{Results}
The predictive results are shown in Table \ref{tab:results_english} (English data sets)\footnote{We only report results of English data sets using OA as those of other languages are not good.} and Table \ref{tab:results_otherLanguage} (data sets in Chinese, Spanish and Italian). Values in the \textit{Performance Drop} column are calculated by subtracting the results of chronological splits from that of random splits, where $\downarrow$ indicates a positive value and $\uparrow$ indicates a negative value. In other words, performance drop refers to the performance decreases using chronological splits compared to random splits with the same model.

\paragraph{Random vs. chronological splits} In general, we observe performance degradation using chronological splits compared to random splits across all pretrained language models (PLMs). This is in line with previous work on other classification tasks such as document classification \citep{chalkidis2022improved}, stance detection \citep{mu2023examining} and rumour detection \citep{mu2023s}. Furthermore, the longer the time span, the greater the performance degradation. For the data sets with long time spans, we observe 16.93$\downarrow$ F1 on WASEEM using RoBERTa and 9.72$\downarrow$ F1 on SANGUINETTI using BERT); while for the data sets with short time spans we observe only 0.5$\downarrow$ F1 on FOUNTA using RoBERTa and 0.77$\downarrow$ F1 on PEREIRA using BERT.

However, although the performance of LR is not as good as that of PLMs, it has a smaller performance drop (or even performance rise) on data sets with small time spans (e.g., 1.26$\uparrow$ F1 on FOUNTA compared with 0.50$\downarrow$ F1 using RoBERTa). 

Interestingly, we observe only a slight performance drop on the data set of JIANG (0.71$\downarrow$ F1 using BERT) despite the eight-year time span. This may be due to the differences in the expression of abusive language online in Chinese and English (JIANG  vs. WASEEM) or different collection methods between these two data sets. Another speculation is that JIANG only focuses on sexist abuse (sexism or not) which is one of the domains of abusive language. In this case, it covers fewer topics than other abusive data sets, which makes the performance less affected by temporalities (we will further investigate it in the following section).

\paragraph{Vanilla vs. domain adaptation models} We compare the vanilla RoBERTa model with the domain adaptation model (RoBERTa-hate-speech) on two English data sets. We found that RoBERTa-hate-speech not only outperforms RoBERTa across two data sets using both random and chronological splits as expected but also has a smaller performance drop on WASEEM  (12.93$\downarrow$), where tweets span three years. This suggests that domain adaptation models can help mitigate temporal bias in abusive language detection, especially over long time spans. However, there are no domain-specific models for other languages, suggesting that further efforts are needed to develop such models.

\paragraph{Zero-Shot Classification} Since OA is trained after the year of WASEEM (2016) and FOUNTA (2018), we hypothesize that the difference of predictive results between two data split strategies using OA will be negligible (e.g. smaller than 1). The performance drop of FOUNTA is as expected (0.23$\uparrow$ F1); while the F1 performance on WASEEM using chronological splits is 7.22 higher than using random splits. We speculate that the large performance difference between these two splitting ways on WASEEM is due to the more explicit abusive content in the testing set using chronological splits as temporalities are less likely to be an influencing factor for OA. To investigate this, we calculate the swearing rates (the percentage of tweets containing at least one swear word among all tweets) of these two testing sets using an English swearword list from Wiktionary (words considered taboo and vulgar or offensive)\footnote{\url{https://en.wiktionary.org/wiki/Category:English_swear_words}}. The swearing rate of WASEEM using random and chronological splits is 5.60\% and 8.40\%; while that of FOUNTA is 4.64\% and 5.51\% respectively. The performance of OA is more likely to be influenced by the explicitness of abusive expressions instead of temporal factors based on the results of two English data sets. However, more abusive data sets are needed to make a more robust conclusion.

\begin{table}[]
\resizebox{0.48\textwidth}{!}{
\begin{tabular}{|l|llllll|}
\hline
            & \multicolumn{3}{c|}{Random Split}                                                        & \multicolumn{3}{c|}{Chronological Split}                                          \\ \hline
            & \multicolumn{1}{l|}{Precision} & \multicolumn{1}{l|}{Recall} & \multicolumn{1}{l|}{F1}   & \multicolumn{1}{l|}{Precision}   & \multicolumn{1}{l|}{Recall}      & F1          \\ \hline
            & \multicolumn{6}{c|}{WASEEM}                                                                                                                                                  \\ \hline
Non-abusive & \multicolumn{1}{l|}{92.6}      & \multicolumn{1}{l|}{91.6}   & \multicolumn{1}{l|}{92.0} & \multicolumn{1}{l|}{77.6 (15.0$\downarrow$)} & \multicolumn{1}{l|}{97.4 (5.8$\uparrow$)}  & 86.6 (5.4$\uparrow$)  \\ \hline
Abusive     & \multicolumn{1}{l|}{82.6}      & \multicolumn{1}{l|}{84.0}   & \multicolumn{1}{l|}{83.0} & \multicolumn{1}{l|}{88.0 (5.4$\uparrow$)}  & \multicolumn{1}{l|}{40.2 (43.8$\downarrow$)} & 55.6 (27.4$\downarrow$) \\ \hline
Overall     & \multicolumn{1}{l|}{87.5}      & \multicolumn{1}{l|}{87.8}   & \multicolumn{1}{l|}{87.6} & \multicolumn{1}{l|}{86.0 (1.5$\downarrow$)}  & \multicolumn{1}{l|}{72.2 (5.6$\downarrow$)}  & 74.7 (12.9$\downarrow$) \\ \hline
            & \multicolumn{6}{c|}{FOUNTA}                                                                                                                                                  \\ \hline
Non-abusive & \multicolumn{1}{l|}{97.6}      & \multicolumn{1}{l|}{98.2}   & \multicolumn{1}{l|}{98.0} & \multicolumn{1}{l|}{96.8 (0.8$\downarrow$)}  & \multicolumn{1}{l|}{98.0 (0.2$\downarrow$)}  & 97.0 (1.0$\downarrow$)  \\ \hline
Abusive     & \multicolumn{1}{l|}{88.6}      & \multicolumn{1}{l|}{83.8}   & \multicolumn{1}{l|}{86.4} & \multicolumn{1}{l|}{86.2 (2.4$\downarrow$)}  & \multicolumn{1}{l|}{77.4 (6.4$\downarrow$)}  & 81.6 (4.8$\downarrow$)  \\ \hline
Overall     & \multicolumn{1}{l|}{93.2}      & \multicolumn{1}{l|}{91.1}   & \multicolumn{1}{l|}{92.1} & \multicolumn{1}{l|}{92.0 (1.2$\downarrow$)}  & \multicolumn{1}{l|}{90.8 (0.3$\downarrow$)}  & 91.4 (0.7$\downarrow$)  \\ \hline
            & \multicolumn{6}{c|}{JIANG}                                                                                                                                                   \\ \hline
Non-abusive & \multicolumn{1}{l|}{83.2}      & \multicolumn{1}{l|}{88.8}   & \multicolumn{1}{l|}{85.8} & \multicolumn{1}{l|}{86.2 (3.0$\uparrow$)}  & \multicolumn{1}{l|}{80.6 (8.2$\downarrow$)}  & 83.2 (2.6$\downarrow$)  \\ \hline
Abusive     & \multicolumn{1}{l|}{74.6}      & \multicolumn{1}{l|}{64.2}   & \multicolumn{1}{l|}{69.0} & \multicolumn{1}{l|}{66.2 (8.4$\downarrow$)}  & \multicolumn{1}{l|}{74.8 (10.6$\uparrow$)} & 70.2 (1.2$\uparrow$)  \\ \hline
Overall     & \multicolumn{1}{l|}{79.0}      & \multicolumn{1}{l|}{76.7}   & \multicolumn{1}{l|}{77.5} & \multicolumn{1}{l|}{76.3 (2.7$\downarrow$)}  & \multicolumn{1}{l|}{77.8 (1.1$\uparrow$)}  & 76.8 (0.7$\downarrow$)  \\ \hline
            & \multicolumn{6}{c|}{PEREIRA}                                                                                                                                                       \\ \hline
Non-abusive & \multicolumn{1}{l|}{85.0}      & \multicolumn{1}{l|}{89.6}   & \multicolumn{1}{l|}{87.4} & \multicolumn{1}{l|}{82.8 (2.2$\downarrow$)}  & \multicolumn{1}{l|}{97.0 (7.4$\uparrow$)}  & 89.2 (1.8$\uparrow$)  \\ \hline
Abusive     & \multicolumn{1}{l|}{65.6}      & \multicolumn{1}{l|}{55.0}   & \multicolumn{1}{l|}{59.6} & \multicolumn{1}{l|}{83.6 (18.0$\uparrow$)}   & \multicolumn{1}{l|}{42.4 (12.6$\downarrow$)} & 55.8 (3.8$\downarrow$)  \\ \hline
Overall     & \multicolumn{1}{l|}{75.3}      & \multicolumn{1}{l|}{72.3}   & \multicolumn{1}{l|}{73.4} & \multicolumn{1}{l|}{83.2 (7.9$\uparrow$)}  & \multicolumn{1}{l|}{69.7 (2.6$\downarrow$)}  & 72.7 (0.7$\downarrow$)  \\ \hline
            & \multicolumn{6}{c|}{SANGUINETTI}                                                                                                                                                       \\ \hline
Non-abusive & \multicolumn{1}{l|}{91.4}      & \multicolumn{1}{l|}{95.0}   & \multicolumn{1}{l|}{93.2} & \multicolumn{1}{l|}{88.2 (3.2$\downarrow$)}  & \multicolumn{1}{l|}{94.6 (0.4$\downarrow$)}  & 91.6 (1.6$\downarrow$)  \\ \hline
Abusive     & \multicolumn{1}{l|}{64.8}      & \multicolumn{1}{l|}{50.4}   & \multicolumn{1}{l|}{56.8} & \multicolumn{1}{l|}{52.2 (12.6$\downarrow$)} & \multicolumn{1}{l|}{31.4 (19.0$\downarrow$)} & 39.0 (17.8$\downarrow$) \\ \hline
Overall     & \multicolumn{1}{l|}{78.1}      & \multicolumn{1}{l|}{72.7}   & \multicolumn{1}{l|}{74.9} & \multicolumn{1}{l|}{70.2 (7.9$\downarrow$)}   & \multicolumn{1}{l|}{63.1 (9.6$\downarrow$)}  & 65.1 (9.8$\downarrow$)  \\ \hline
\end{tabular}}
\caption{Model predictive performance of each class as well as the overall performance using random and chronological splits.}
\label{tab:results_per_class}
\end{table}

We further explore whether temporal bias has a greater influence on abusive texts or non-abusive texts. Table \ref{tab:results_per_class} shows the performance of each class as well as the overall performance on five data sets using their best-performing models (\textit{RoBERTa-hate-speech} for English data sets and \textit{BERT} for other language data sets). In general, the performance drop in abusive classes is larger than that in non-abusive classes. Also, the larger the time span of the data sets, the greater the difference in performance degradation between abusive and non-abusive classes (e.g. F1 1.8$\uparrow$ vs. 27.4$\downarrow$ for PEREIRA with ten-month time span and F1 1.6$\downarrow$ vs. 17.8$\downarrow$ for SANGUINETTI with two-year time span). However, \citet{jiang2022swsr} is an exception where F1 scores of abusive classes increase by 1.2. We also notice that the degradation of precision for non-abusive content is larger than that of recall using chronological splits (e.g. 3.2$\downarrow$ precision and 0.4$\downarrow$ recall in SANGUINETTI); while for abusive content, the performance drop in precision and recall is reversed (e.g. 5.4$\uparrow$ precision and 43.8$\downarrow$ in recall in WASEEM). This indicates that by using chronological splits, non-abusive texts are more likely to be detected; fewer abusive texts can be detected but the detected ones are more likely to be correct.

\begin{table}[]
\resizebox{0.48\textwidth}{!}{
\begin{tabular}{|l|l|l|l|l|}
\hline
\textbf{Data Set} & \textbf{Splits} & \textbf{Jarccard} & \textbf{DICE} & \textbf{OC} \\ \hline
\multirow{3}{*}{\textbf{WASEEM}} & \textit{Random} & .278 & .435 & .777 \\ \cline{2-5} 
& \textit{Chronological} & .216 & .355 & .733 \\ \cline{2-5}
& \cellcolor{mGray} Similarity Drop & \cellcolor{mGray} .062$\downarrow$ & \cellcolor{mGray} .080$\downarrow$ & \cellcolor{mGray} .044$\downarrow$ \\
\hline
\multirow{3}{*}{\textbf{FOUNTA}} & Random & .203 & .337 & .672   \\ \cline{2-5} 
& \textit{Chronological} & .199 & .332 & .668 \\ \cline{2-5}
& \cellcolor{mGray} Similarity Drop & \cellcolor{mGray} .004$\downarrow$ & \cellcolor{mGray} .005$\downarrow$ & \cellcolor{mGray} .004$\downarrow$ \\
\hline
\multirow{3}{*}{\textbf{JIANG}} & \textit{Random} & .243 & .391 & .748 \\ \cline{2-5} 
& Chronological & .211 & .349 & .717 \\ \cline{2-5}
& \cellcolor{mGray} Similarity Drop & \cellcolor{mGray} .032$\downarrow$ & \cellcolor{mGray} .042$\downarrow$ & \cellcolor{mGray} .031$\downarrow$\\
\hline
\multirow{3}{*}{\textbf{PEREIRA}} & \textit{Random} & .185 & .312 & .653 \\ \cline{2-5} 
& \textit{Chronological} & .167 & .286 & .602 \\  \cline{2-5}
& \cellcolor{mGray} Similarity Drop & \cellcolor{mGray} .018$\downarrow$ & \cellcolor{mGray} .026$\downarrow$ & \cellcolor{mGray} .051$\downarrow$\\
\hline
\multirow{3}{*}{\textbf{SANGUINETTI}} & \textit{Random} & .190 & .320 & .657 \\ \cline{2-5} 
& \textit{Chronological} & .173 & .295 & .636 \\ \cline{2-5}
& \cellcolor{mGray} Similarity Drop & \cellcolor{mGray} .017$\downarrow$ & \cellcolor{mGray} .025$\downarrow$ & \cellcolor{mGray} .021$\downarrow$\\
\hline
\end{tabular}}
\caption{Text similarities between training and testing sets using Jarccard, DICE and OC.}
\label{tab:similarity}
\end{table}

\section{Analysis}
\label{sec:analysis}

\subsection{Text Similarities}
We hypothesize that the drop in performance is due to a larger difference between training and testing sets using chronological splits. To verify this, we use three methods to calculate text similarities: (a) Jaccard similarity coefficient; (b) DICE coefficient \citep{dice1945measures} and (c) overlap coefficient (OC).

\textbf{Jaccard similarity coefficient} is defined as the size of the intersection divided by the size of the union of two sets, A and B,
\begin{equation}
    J(A,B) = \frac{|A \cap B|}{|A \cup B|}
\end{equation}

\textbf{DICE cofficient} is defined as twice the size of the intersection divided by the sum size of two sets, A and B,
\begin{equation}
    DICE(A,B) = \frac{2 * |A \cap B|}{|A| + |B|}
\end{equation}

\textbf{Overlap coefficient} is defined as the size of the intersection divided by the smaller size of the two sets, A and B,
\begin{equation}
    OC(A,B) = \frac{|A \cap B|}{\min(|A|, |B|)}
\end{equation}

where A and B denote the set of distinctive words from training and test sets, respectively. $|A \cap B|$
and $|A \cup B|$  indicate the sum of distinctive words that appear in the intersection and union of the two subsets respectively. When the two subsets have no shared vocabulary, these three coefficient values will be zero, while if they are identical, the two values will be 1.

Table \ref{tab:similarity} shows the similarity coefficient between training and testing sets using \textit{random} and \textit{chronological} splits. Firstly, we notice that values from three similarity calculation methods drop across all data sets, indicating that using chronological splits leads to a larger difference between training and testing sets. Secondly, the longer the time span of data sets, the larger the similarity drop. For example, OC of WASEEM (three years) drops 0.044 while that of FOUNTA (one week) drops 0.004. Also, there tends to be a positive correlation between the magnitude of similarity reduction and the performance drop. However, considering the minor decline (drop 0.71 F1) in the predictive performance of JIANG (eight years), the text similarity drop is not consistent (e.g. OC drops 0.31). This can be explained by the fact that text similarity calculation is granular down to words; while topics might be limited (number, variety) in a sexist data set (i.e. JIANG).

\renewcommand{\arraystretch}{1.2}
\begin{table}[!t]
\center
\resizebox{0.48\textwidth}{!}{
\begin{tabular}{|l|c|l|c||l|c|l|c|}
\hline
\multicolumn{4}{|c||}{\bf Random Splits} & \multicolumn{4}{|c|}{\bf Chronological Splits} \\
\hline
\multicolumn{2}{|c|}{\bf Training} & \multicolumn{2}{|c||}{\bf Testing} & \multicolumn{2}{|c|}{\bf Training} & \multicolumn{2}{|c|}{\bf Testing}\\
\hline
Unigram & r & Unigram & r & Unigram & r & Unigram & r \\ 
\hline
mohammed & .030 & countless & .056 & sexist & .163 & kat & .550\\
\hline
liar & .029 & chipotle & .056 & women & .116 & \#mkr & .459\\
\hline
\#mkr2015 & .028 & rapes & .056 & islam & .089 & andre & .230\\
\hline
job & .028 & fault & .054 & \#notsexist & .078 & face & .165\\
\hline
day & .027 & lower & .050 & call & .071 & annie & .161 \\
\hline
truth & .026 & distraction & .049 & female & .070 & \#cuntandandre & .147 \\
\hline
kat & .026 & forget & .047 & men & .065 & \#katandandre & .121\\
\hline
everything & .026 & terrorist & .047 & girls & .060 & celine & .112\\
\hline
death & .025 & consider & .047 & religion & .051 & karma & .109\\
\hline
fight & .023 & appears & .047 & prophet & .049 & cunt & .099\\
\hline
\end{tabular}}
\caption{Unigram feature correlations with abusive tweets between training and testing sets from WASEEM using random splits (left) and chronological splits (right), sorted by Pearson correlation (r). All correlations are significant at $p < .001$, two-tailed t-test.}
\label{t:feature_en}
\end{table}


\renewcommand{\arraystretch}{1.2}

\begin{table}[!t]
\center
\resizebox{0.48\textwidth}{!}{
\begin{tabular}{|l|c|l|c|}
\hline
\multicolumn{2}{|c|}{\bf Training} & \multicolumn{2}{|c|}{\bf Testing}\\
\hline
\multicolumn{4}{|c|}{\bf Random Splits} \\

\hline
Unigram & r & Unigram & r \\ 
\hline

\begin{CJK*}{UTF8}{gbsn}
性别
\end{CJK*} (gender) & .053 & 
\begin{CJK*}{UTF8}{gbsn}
压迫者
\end{CJK*} (oppressor) & .084\\
\hline
\begin{CJK*}{UTF8}{gbsn}
小
\end{CJK*} (small) & .044 & 
\begin{CJK*}{UTF8}{gbsn}
多点
\end{CJK*} (more) & .083\\
\hline
\begin{CJK*}{UTF8}{gbsn}
叫
\end{CJK*} (shout) & .040 & 
\begin{CJK*}{UTF8}{gbsn}
取决于
\end{CJK*} (depending on) & .074\\
\hline
\begin{CJK*}{UTF8}{gbsn}
搞
\end{CJK*} (do) & .039 & 
\begin{CJK*}{UTF8}{gbsn}
并非
\end{CJK*} (not) & .073\\
\hline
\begin{CJK*}{UTF8}{gbsn}
样子
\end{CJK*} (looks) & .034 & 
\begin{CJK*}{UTF8}{gbsn}
发出
\end{CJK*} (sending) & .072\\
\hline
\begin{CJK*}{UTF8}{gbsn}
先
\end{CJK*} (first) & .033 & 
\begin{CJK*}{UTF8}{gbsn}
废话
\end{CJK*} (nonsense) & .072\\
\hline
\begin{CJK*}{UTF8}{gbsn}
之前
\end{CJK*} (before) & .033 & 
\begin{CJK*}{UTF8}{gbsn}
承担责任
\end{CJK*} (take responsibility) & .071\\
\hline
\multicolumn{4}{|c|}{\bf Chronological Splits} \\

\hline
\begin{CJK*}{UTF8}{gbsn}
人
\end{CJK*} (people) & .063 & 
\begin{CJK*}{UTF8}{gbsn}
厌女
\end{CJK*} (misogyny) & .132\\
\hline
\begin{CJK*}{UTF8}{gbsn}
多
\end{CJK*} (more) & .051 & 
\begin{CJK*}{UTF8}{gbsn}
厌
\end{CJK*} (hate) & .122\\
\hline
\begin{CJK*}{UTF8}{gbsn}
那么
\end{CJK*} (then) & .042 & 
\begin{CJK*}{UTF8}{gbsn}
女上司
\end{CJK*} (female manager) & .098\\
\hline
\begin{CJK*}{UTF8}{gbsn}
人家
\end{CJK*} (people) & .042 & 
\begin{CJK*}{UTF8}{gbsn}
下属
\end{CJK*} (subordinate) & .092\\
\hline
\begin{CJK*}{UTF8}{gbsn}
需要
\end{CJK*} (need) & .040 & 1 & .088 \\
\hline
\begin{CJK*}{UTF8}{gbsn}
或者
\end{CJK*} (or) & .040 &
\begin{CJK*}{UTF8}{gbsn}
介意
\end{CJK*} (mind) & .079 \\
\hline
\begin{CJK*}{UTF8}{gbsn}
只是
\end{CJK*} (just) & .0.039 & 
\begin{CJK*}{UTF8}{gbsn}
我
\end{CJK*} (I) & .079\\
\hline
\end{tabular}}
\caption{Unigram feature correlations with abusive tweets between training and testing sets from JIANG using random splits (left) and chronological splits (right), sorted by Pearson correlation (r). All correlations are significant at $p < .001$, two-tailed t-test.}
\label{t:feature_ch}
\end{table}

\begin{figure*}[t!]
     \centering
     \begin{subfigure}[b]{0.48\textwidth}
         \centering
         \includegraphics[width=\textwidth]{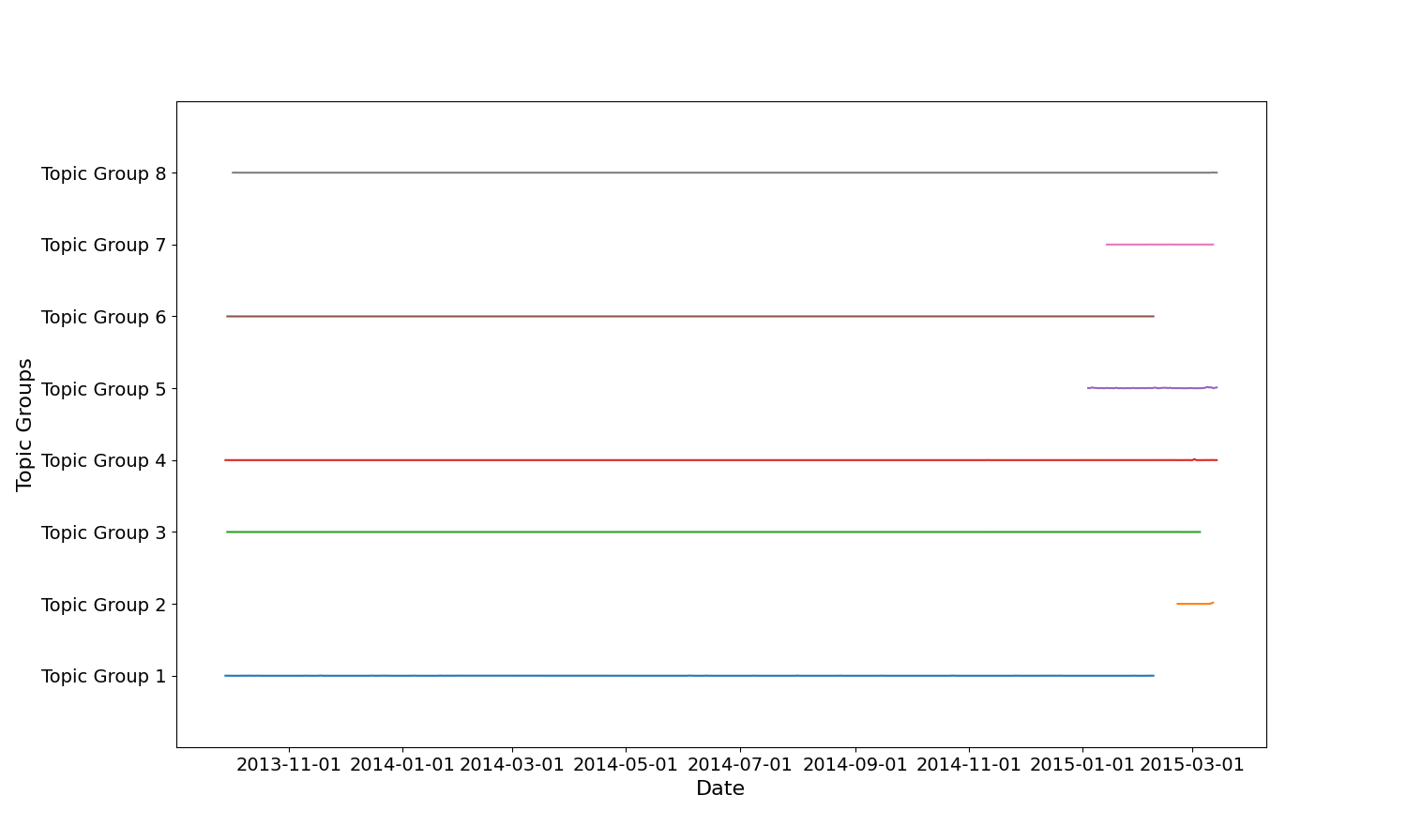}
         \caption{Topic distribution of WASEEM.}
         \label{fig:waseem_topic}
     \end{subfigure}
     \hfill
     \begin{subfigure}[b]{0.48\textwidth}
         \centering
         \includegraphics[width=\textwidth]{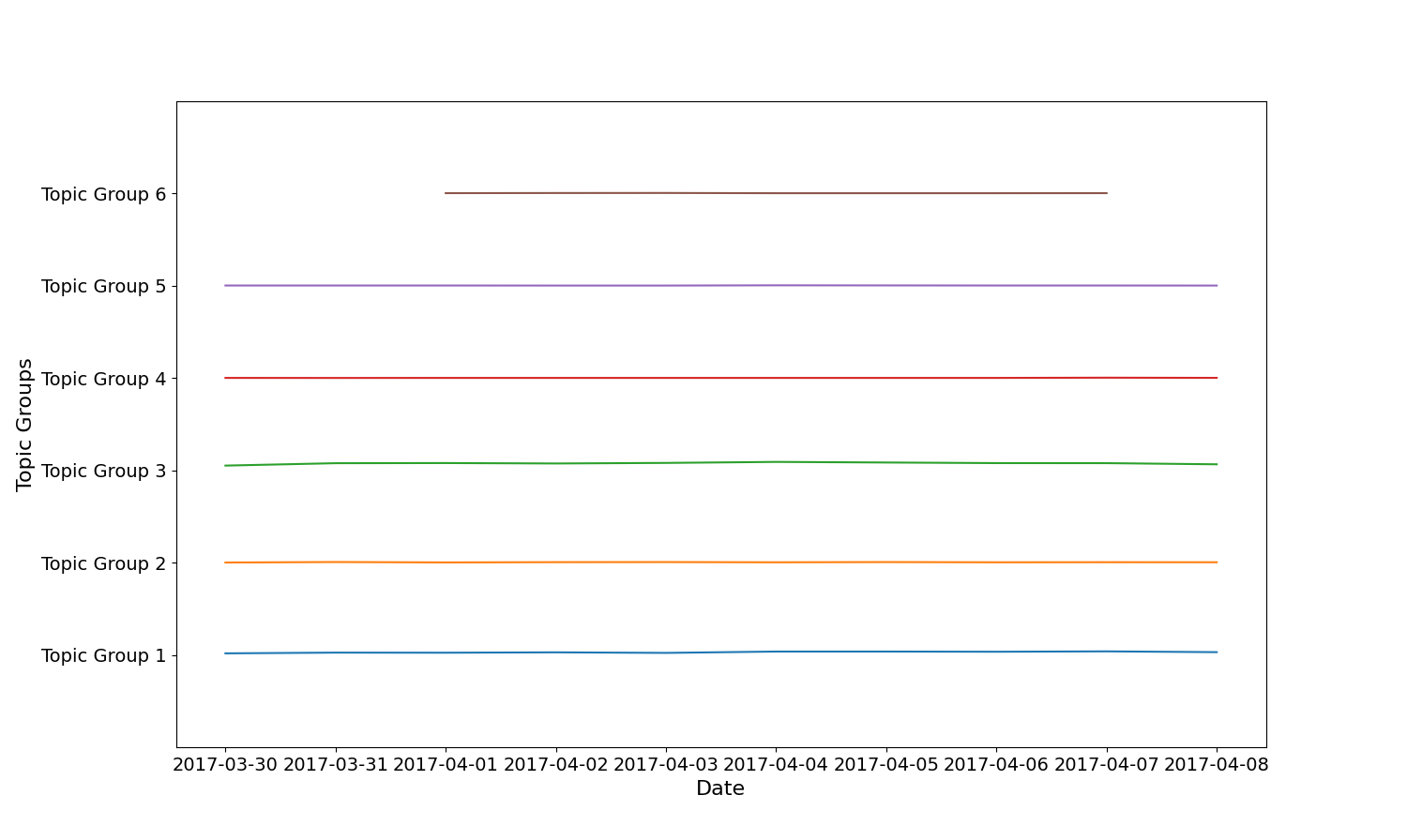}
         \caption{Topic distribution of FOUNTA.}
         \label{fig:founta_topic}
     \end{subfigure}
        \caption{Topic distribution over time.}
        \label{fig:topic_distribution}
\end{figure*}

\subsection{Linguistic Analysis}
We hypothesize that when using chronological splits, there are more new events or topics in the testing set that are not present in the training set, which leads to a decrease in model performance. In contrast, when using random splits, topics are evenly distributed between the training and testing sets. We present the linguistic patterns (unigram) of abusive tweets in training and testing data sets using two splitting strategies involving univariate Pearson correlations. Table \ref{t:feature_en} shows unigram feature correlations of WASEEM and Table \ref{t:feature_ch} shows that of JIANG. We compare these two data sets because their time spans are both long (three years vs. eight years) while their predictive performance drops vary widely (16.93 vs. 0.71).

For WASEEM, most abusive tweets in the testing set using chronological splits involve an Australian TV show, My Kitchen Rules (MKR) (e.g. \textit{\#mkr}, \textit{\#cuntandandre}, \textit{\#kateandandre}, \textit{kat}, \textit{andre}, \textit{annie}), which is one of the queried terms for data collection. Our speculation is that the discussion about this show began to emerge during the later timeframe of the data set (within the time covered by the testing set when using chronological splits). However, there are hardly any new topics in the testing set when using random splits (e.g. \textit{countless}, \textit{lower}, \textit{forget}).

For JIANG, testing sets using both split strategies mainly contain basic or gender-related terms (e.g. \textit{more}, \textit{not}, \textit{misogyny}, \textit{female manager}) and do not involve terms related to specific events. This is also correlated to how they collect the data: searching gender-related keywords such as `feminism' and `gender discrimination' for sexist content instead of using specific events as keywords. This suggests that collecting data using generic terms as keywords instead of terms associated with current hot events is likely to introduce less temporal bias.

\subsection{Topic Distribution}
We also explore topic distribution over time across two English data sets. We first use a topic modelling technique, BERTopic\footnote{https://github.com/MaartenGr/BERTopic}, to extract the 10 most important topic groups in a data set. Then we manually remove repeated or commonly used words (e.g. `this', `said') in these topic groups and combine similar groups into one group (e.g. combining `women', `men', `she', and `girls' into \textit{gender-related} group). The generated topic groups of each data set are shown as follows\footnote{We also try to extract topics of data sets with other language using BERTopic but the results are not good.}:

\paragraph{WASEEM:} Group 1: \{\textit{sexist}, \textit{women}, \textit{men}, \textit{bitch}, \textit{her}, \textit{she}, \textit{girls}, \textit{female}, \textit{woman}, \textit{notsexist}\}; Group 2: \{\textit{kat}, \textit{mkr}, \textit{face}, \textit{mkr2015}, \textit{karma}\}; Group 3: \{\textit{drive}, \textit{drivers}, \textit{driving}, \textit{driver}\}; Group 4: \{\textit{blondes}, \textit{blonde}, \textit{pretty}, \textit{hot}, \textit{dumb}\}; Group 5: \{\textit{israel}, \textit{hamas}, \textit{palestinians}, \textit{israelis}, \textit{palestinian}, \textit{palestine}, \textit{gays}, \textit{destroy}, \textit{muslims}\}; Group 6: \{\textit{sports}, \textit{announcers}, \textit{commentators}, \textit{announcer}, \textit{football}, \textit{stand}, \textit{commentator}\}; Group 7: \{\textit{feminism}, \textit{feminists}, \textit{feminist}, \textit{equality}, \textit{movement}, \textit{hypocrisy}, \textit{rights}, \textit{emma}, \textit{modern}\}; Group 8: \{\textit{funny}, \textit{comedians}, \textit{comedian}, \textit{jokes}\}.

\paragraph{FOUNTA:} Group 1: \{\textit{trump}, \textit{president}, \textit{obama}, \textit{voted}, \textit{republicans}, \textit{idiot}\}; Group 2: \{\textit{nigga}, \textit{niggas}\}; Group 3: \{\textit{hate}, \textit{bitch}, \textit{bad}, \textit{fucking}, \textit{bitches}, \textit{she}\}; Group 4: \{\textit{syria}, \textit{assad}, \textit{syrian}, \textit{chemical}, \textit{trump}, \textit{missiles}, \textit{attack}, \textit{obama}, \textit{war}, \textit{refugees}\}; Group 5: \{\textit{pizza}, \textit{eat}, \textit{pineapple}, \textit{disgusting}, \textit{food}, \textit{home}, \textit{taco}\}; Group 6: \{\textit{wrestlemania}, \textit{wwe}, \textit{match}, \textit{rawaftermania}, \textit{wrestlemania33}\}.

Figure \ref{fig:topic_distribution} shows the topic distributions over time of these two data sets. For WASEEM,  Group 2 (MKR TV show related ), 5 (race and religion related) and 7 (feminism related) appear only after 2015, which is also the starting time of the testing data set using chronological splits (March 2015). This results in the models barely seeing these words in the training set and a lack of knowledge in these three topics during training, especially for Group 2. Thus, it would be easier for models to fail when predicting text involving these topics using chronological splits. All topic groups are evenly distributed in FOUNTA except for Group 6 (wrestling match related). However, Topic Group 6 rarely appears in the testing set using chronological splits (starting from 7th April 2017), which is less likely to influence the performance.

\subsection{Filtered Data Set}

We explore whether removing words related to specific topics or events will enhance the robustness of the models when predicting abusive content. We hypothesize that the model performance will drop slightly while the difference between random and chronological splits will be more minor by removing these words. We experiment with WASEEM as its performance drop has room to reduce. We filter the data set by excluding three types of words: (1) words in all eight groups extracted by BERTopic (\textbf{D1}); (2) words selected by attention mechanisms (\textbf{D2}) and (3) the union of the words extracted by (1) and (2) (\textbf{D3)}. For (2), we first use the RoBERTa-hate-speech model to produce attention scores that represent a probability distribution over each text. We then manually remove topic-related tokens among the top five tokens with the highest probability in each abusive tweet. Most of the removed tokens are names or hashtags related to the cooking TV show.

The results of filtered data sets are shown in Table \ref{tab:results_filtered}. Similar to the previous experiment, we run five times for each method. First, all three strategies for removing topic-related words hurt performance in most cases, especially for chronological splits (e.g. 87.64 vs. 84.75 F1 using random splits, 74.71 vs. 72.11 F1 using chronological splits). However, the performance on D2 using chronological splits improves by 0.32 F1. Second, using more robust data sets leads to more minor performance drops. We achieve the smallest performance drop (9.65$\downarrow$ F1) using D3. Also, using D2 achieves a comparable performance drop but only slightly hurts the performance. This suggests that filtering out specific topic-related words in a data set (i.e. a more robust data set) helps reduce temporal bias.

\begin{table}[]
\begin{tabular}{|l|llll|}
\hline
                 & \multicolumn{1}{l|}{Acc}   & \multicolumn{1}{l|}{P}     & \multicolumn{1}{l|}{R}     & F1    \\ \hline
                 & \multicolumn{4}{l|}{\textbf{Without removal}}                                                         \\ \hline
\textit{Random}           & \multicolumn{1}{l|}{89.20} & \multicolumn{1}{l|}{87.50} & \multicolumn{1}{l|}{87.82} & 87.64 \\ \hline
\textit{Chronological}    & \multicolumn{1}{l|}{81.58} & \multicolumn{1}{l|}{85.99} & \multicolumn{1}{l|}{72.21} & 74.71 \\ \hline
\rowcolor{mGray} Performance drop & \multicolumn{1}{l|}{7.62$\downarrow$}  & \multicolumn{1}{l|}{1.51$\downarrow$}  & \multicolumn{1}{l|}{15.61$\downarrow$} & 12.93$\downarrow$ \\ \hline
                 & \multicolumn{4}{l|}{\textbf{D1: Remove words by BERTopic}}                                                \\ \hline
\textit{Random}           & \multicolumn{1}{l|}{86.96} & \multicolumn{1}{l|}{85.42} & \multicolumn{1}{l|}{84.20} & 84.75 \\ \hline
\textit{Chronological}    & \multicolumn{1}{l|}{80.45} & \multicolumn{1}{l|}{83.86} & \multicolumn{1}{l|}{70.93} & 73.21 \\ \hline
\rowcolor{mGray} Performance drop & \multicolumn{1}{l|}{6.51$\downarrow$}  & \multicolumn{1}{l|}{1.56$\downarrow$}  & \multicolumn{1}{l|}{13.27$\downarrow$} & 11.54$\downarrow$ \\ \hline
                 & \multicolumn{4}{l|}{\textbf{D2: Remove words by attention}}                                               \\ \hline
\textit{Random}           & \multicolumn{1}{l|}{87.16} & \multicolumn{1}{l|}{85.75} & \multicolumn{1}{l|}{84.30} & 84.95 \\ \hline
\textit{Chronological}    & \multicolumn{1}{l|}{81.50} & \multicolumn{1}{l|}{84.61} & \multicolumn{1}{l|}{72.61} & 75.03 \\ \hline
\rowcolor{mGray} Performance drop & \multicolumn{1}{l|}{5.66$\downarrow$}  & \multicolumn{1}{l|}{\textbf{1.14}$\downarrow$}  & \multicolumn{1}{l|}{11.69$\downarrow$} & 9.92$\downarrow$  \\ \hline
                 & \multicolumn{4}{l|}{\textbf{D3: Remove words by both}}                                                             \\ \hline
\textit{Random}           & \multicolumn{1}{l|}{84.73} & \multicolumn{1}{l|}{83.37} & \multicolumn{1}{l|}{80.70} & 81.76 \\ \hline
\textit{Chronological}    & \multicolumn{1}{l|}{79.35} & \multicolumn{1}{l|}{80.89} & \multicolumn{1}{l|}{70.10} & 72.11 \\ \hline
\rowcolor{mGray} Performance drop & \multicolumn{1}{l|}{\textbf{5.38}$\downarrow$}  & \multicolumn{1}{l|}{2.48$\downarrow$}  & \multicolumn{1}{l|}{\textbf{10.60}$\downarrow$} & \textbf{9.65}$\downarrow$  \\ \hline
\end{tabular}
\caption{Model predictive performance using RoBERTa-hate-speech on WASEEM with and without filtering. The smallest performance drops across filtering strategies are in bold.}
\label{tab:results_filtered}
\end{table}

\subsection{Error Analysis}
Additionally, we perform an error analysis on two data sets containing sexist abuse, WASEEM and JIANG, using chronological splits. For WASEEM, we found that most errors happen when content involves the TV show (MKR). Also, when names from the show are mentioned, it is easy for models to misclassify the texts as non-abusive. We guess this is because the model cannot associate names in the testing set with male, female (gender-related) or abusive if it has not seen those names in the training set. However, the annotators of this data set have prior knowledge of this TV show and its characters. Thus, they are able to classify dissatisfaction or hatred toward specific characters as \textit{sexist}. In the following two examples, tweets belonging to \textit{abusive} are misclassified as \textit{non-abusive} (names are highlighted in bold)\footnote{Note that WASEEM is originally a sexist and racist data set, so other abusive content will be labeled as neither (\textit{non-abusive} in our paper).}:

\begin{quote}
    T1: \textit{\textbf{Kat} on \#mkr is such a horrible person.. I wish \textbf{Kat} and \textbf{Andre} would just get eliminated already.}
\end{quote}
\begin{quote}
    T2: \textit{\#MKR-I am seriously considering not watching just because I have to see \textbf{Kats} face. God. I want to slap it with a spatula!}
\end{quote}
However, when gender-related words also appear in the content, models are more likely to classify them correctly. The following
tweets are correctly classified as \textit{abusive}:

\begin{quote}
    T3: \textit{\#katandandre gaaaaah I just want to slap \textbf{her} back to WA \#MKR}
\end{quote}

\begin{quote}
    T4: \textit{\#MKR \textbf{Girls}, thank you for filling the slapper quotient on this years series... we no longer have a need for \textbf{bitchy blondes}! Au Revoir!}
\end{quote}

For JIANG, it is easy for models to fail to understand the actual meaning of a text without knowing traditional Chinese cultural viewpoints related to gender and marriage (e.g. some people value sons more than daughters). The following text belong to \textit{abusive} (sexism originally) is wrongly classified as \textit{non-abusive}:
\begin{quote}
    T5: \begin{CJK*}{UTF8}{gbsn}
\textit{什么垃圾父亲，大女儿16岁就嫁人生子，没达到法定结婚年龄真的没问题？拿法律规定是用来干嘛的？又接着逼15岁二女儿去相亲赚彩礼？养猪啊？尽早出栏降低自己的 成本是吗？}
\end{CJK*} (\textit{What a terrible father! Marrying off his eldest daughter and letting her have a child at the age of 16, without meeting the legal marriage age requirement? What's the point of having laws if they're not followed? And now he's pressuring his 15-year-old second daughter to go on blind dates to earn a dowry? Is he treating them like livestock? Trying to reduce his own costs by selling them off early?})
\end{quote}
Furthermore, objective discussions that contain words closely related to abuse are more likely to be misclassified as \textit{abusive}. The following text is an example (potential abusive words are in bold):
\begin{quote}
    T6: \begin{CJK*}{UTF8}{gbsn}
\textit{家暴不分男女！精神和身体上的暴力是同等的！}
\end{CJK*} (\textit{\textbf{Domestic violence} knows no gender! \textbf{Mental and physical violence} are equally harmful!})
\end{quote}

To conclude, for WASEEM, models tend to misclassify tweets containing terms that implicitly link to gender or sexist where models have no prerequisite knowledge; while for JIANG, most errors happen when involving Chinese culture or terms that are more likely to appear in abusive content.

\section{Limitations}
This work aims to investigate the impact and causes of temporalities across different abusive data sets. In our work, we can only evaluate limited data sets that provide time information (e.g. 2 English ones, 2 data sets spanning more than 3 years) which limits control experiments for more sound comparisons. Also, all debiasing methods can only applied to English abusive data sets due to the imperfect implementation of techniques in other languages (i.e. domain adaptation models, BERTopic, OA). Moreover, our studies on temporal bias only explore topic changes and lack a comprehensive understanding of language evolution over time.

\section{Conclusion}
In this work, we investigate the impact of temporal bias on abusive language detection. We compare the predictive results using two data split methods (i.e. random and chronological splits) across different data sets (\textit{RQ1}). The results indicate that temporal bias has a larger influence on data sets with a larger time span and collected using keywords, especially specific event-related keywords. Languages (or culture) may also be a factor but due to insufficient data sets, we can not draw concrete conclusions. We also conduct extensive analysis including text similarities, feature analysis and topic distribution to explore the causes of temporalities (\textit{RQ2}). We found that performance degradation is mostly because of topic changes in our data sets. To provide a complete answer to \textit{RQ3}, we filter a data set by removing topic-related words that appear in abusive texts. The predictive results suggest that using domain adaptation models and LLMs and training on a more robust data set can effectively reduce temporal bias in abusive language detection.

In the future, we plan to study temporal bias patterns in abusive data sets across different languages or platforms, aiming to understand the importance of considering the specific nature of the target variable when collecting the data sets and developing models. It can also be expanded to other text classification tasks.

\section{Ethics Statement}
This work has received ethical approval from our Research Ethics Committee. All datasets are acquired either through the URLs provided in the original papers or by requesting them from the respective authors. Note that we did not gather any fresh data from Twitter for this study. Additionally, we can verify that the data has been completely anonymized prior to its utilization in the Language Model Inference process. 

\section{Acknowledgements}
This research is supported by a UKRI grant ES/T012714/1 (“Responsible AI for Inclusive, Democratic Societies: A cross-disciplinary approach to detecting and countering abusive language online”).

\bibliography{aaai22}

\end{document}